\pdfoutput=1
\documentclass[11pt]{article}

\usepackage{acl}

\usepackage[ruled,vlined]{algorithm2e}
\usepackage{booktabs}
\usepackage{comment}

\usepackage{times}
\usepackage{latexsym}
\usepackage{amssymb}
\usepackage{amsmath}
\usepackage[T1]{fontenc}
\usepackage[utf8]{inputenc}

\usepackage{microtype}

\usepackage{inconsolata}

\usepackage{graphicx}

\usepackage{enumitem}

\title{TNG-CLIP:\\Training-Time Negation Data Generation for Negation Awareness of CLIP}

\author{Yuliang Cai \thanks{\ The github repository of the code can be found \href{https://github.com/YuliangCai2022/TNG-CLIP}{here}.}\\
  University of Southern California \\
  \texttt{caiyulia@usc.edu} \\\And
 Jesse Thomason \\
  University of Southern California\\
  \texttt{jessetho@usc.edu} 
  \AND Mohammad Rostami \\
  University of Southern California \\
  \texttt{rostamim@usc.edu}   }

\begin{document}
\maketitle
\begin{abstract}
Vision-language models (VLMs), such as CLIP, have demonstrated strong performance across a range of downstream tasks. However, CLIP is still limited in negation understanding: the ability to recognize the absence or exclusion of a concept. Existing methods address the problem by using a large language model (LLM) to generate large-scale data of image captions containing negation for further fine-tuning CLIP. However, these methods are both time- and compute-intensive, and their evaluations are typically restricted to image-text matching tasks. To expand the horizon, we (1) introduce a training-time negation data generation pipeline such that negation captions are generated during the training stage, which only increases 2.5\% extra training time, and (2) we propose the first benchmark, \textsc{Neg-TtoI}, for evaluating text-to-image generation models on prompts containing negation, assessing model's ability to produce semantically accurate images. We show that our proposed method, \textit{TNG-CLIP}, achieves SOTA performance on diverse negation benchmarks of image-to-text matching, text-to-image retrieval, and image generation.

\end{abstract}

\section{Introduction}
Vision-language models (VLM), such as CLIP \citep{radford2021learningtransferablevisualmodels}, provide an efficient approach to tackle vision-language tasks by learning the features of different modalities in a shared embedding space. However, these models fundamentally lack a robust understanding of \textbf{negation}---the ability to recognize  the absence or exclusion of a concept, \textit{e.g., ``A dog \textbf{not} playing a ball.'', ``There is \textbf{no} tree on the street.''}. Negation is a fundamental  aspect of human reasoning, enabling precise description of constraints and expectations in communication. Without proper negation understanding, VLMs generate and retrieve semantically incorrect content, particularly in complicated scenarios where the presence or absence of specific elements critically alters meanings. 

\begin{figure}
\centering
\includegraphics[scale=0.4]{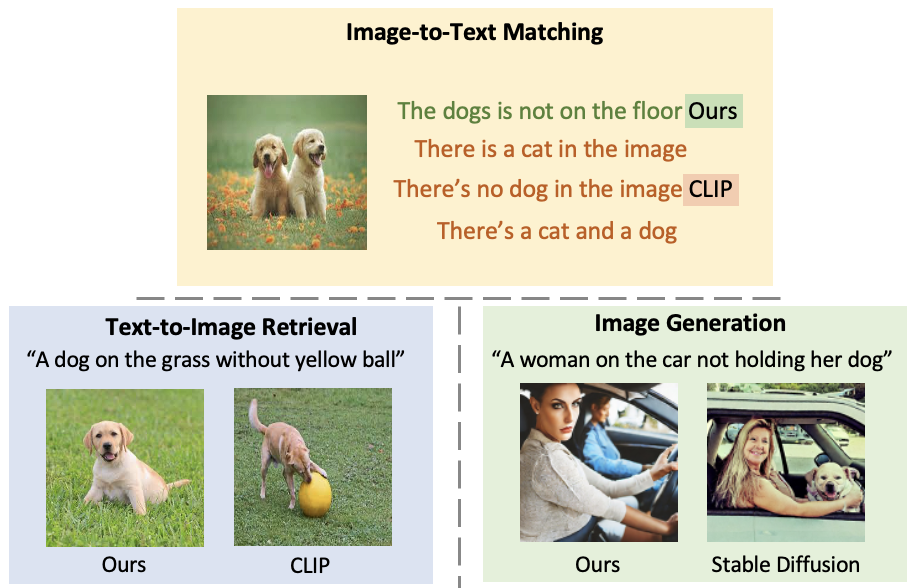}
\caption{We present \textit{TNG-CLIP}, a negation-aware CLIP that achieves outstanding negation understanding in image-to-text matching, text-to-image retrieval and proposed image generation \textsc{Neg-TtoI} benchmarks.}
\label{fig:first}
\end{figure}

To tackle this problem, current methods \citep{negbench,conclip,park2025knownobetterdatadriven,negclip} focus on generating well-designed image-text datasets, such that there are negation captions associated with each image sample, and then fine-tune the underlying VLM. However, such approaches face three challenges: (1) the negation of each caption is designed, generated, and verified via LLMs. Considering the fact that the existing vision-language datasets \citep{chen2015microsoftcococaptionsdata,changpinyo2021conceptual12mpushingwebscale} contain millions of samples, generating the negation dataset is extremely time- and compute-consuming. (2) Unlike standard semantic descriptions, which are typically grounded in observable features, the negation process introduces arbitrariness by specifying the absence of concepts that are not depicted. For example, given an image of \textit{``a dog playing a ball''}, one could construct multiple valid negation  captions such as \textit{``a dog playing a ball while no man is present''} or \textit{``a dog playing a ball but not on the beach''}. By generating fixed negation captions, previous methods may constrain the diversity of negation scenarios, thus harming the   generalization of the fine-tuned VLM on negation understanding tasks. (3) Previous methods are mainly evaluated on image-to-text matching and text-to-image retrieval tasks. Considering the versatility of CLIP, however, evaluation should not be constrained to matching-based tasks and must include more diverse downstream tasks such as generation-based tasks, where the text encoder can be used as part of a generative model \citep{rombach2022highresolutionimagesynthesislatent}.

We propose a new data generation and training pipeline which generates negation captions during  training without the need for a pre-defined negated image-text pair dataset. In each training batch, we identify the most similar image–text pair for every image–text example by computing the cosine similarity between their embedded image features. For each caption, we generate negated variants using a template-based approach, by interacting with another caption in the same batch. Because the negated caption generation relies on the other captions, we can generate diverse and different negated captions in every training epoch. We also propose a negation text-to-image generation benchmark, \textsc{Neg-TtoI}, to evaluate the capability of models to avoid generating undesired objects given negated prompts. In this task, a compositional negated caption is given which contains the desired objects and undesired objects, \textit{e.g., ``A women not holding a dog in the car''}. The generative model needs to explicitly recognize what needs to be generated and what should be avoided. We show that our proposed data generation and training pipeline   can directly benefit the downstream task of text-to-image generation. Our   contributions include:
\begin{itemize}[noitemsep,nosep]
    \item  We propose a novel and efficient training-time negation generation pipeline, \textit{TNG-CLIP}, to improve CLIP's negation understanding by generating dynamic and diverse negation samples during training without the need for LLMs and pre-defined negation datasets. 
    \item  We propose the first benchmark for negation-aware text-to-image generation task, \textsc{Neg-TtoI}, which contains diverse and abundant samples to evaluate model's negation understanding capability.
    \item  We offer extensive experiments to demonstrate that  \textit{TNG-CLIP} achieves SOTA performance on diverse negation-aware downstream tasks including image-to-text matching, text-to-image retrieval, and image generation, indicating its robustness across these tasks.
\end{itemize}

\section{Related Works}
\begin{figure*}
\centering
\includegraphics[scale=0.49]{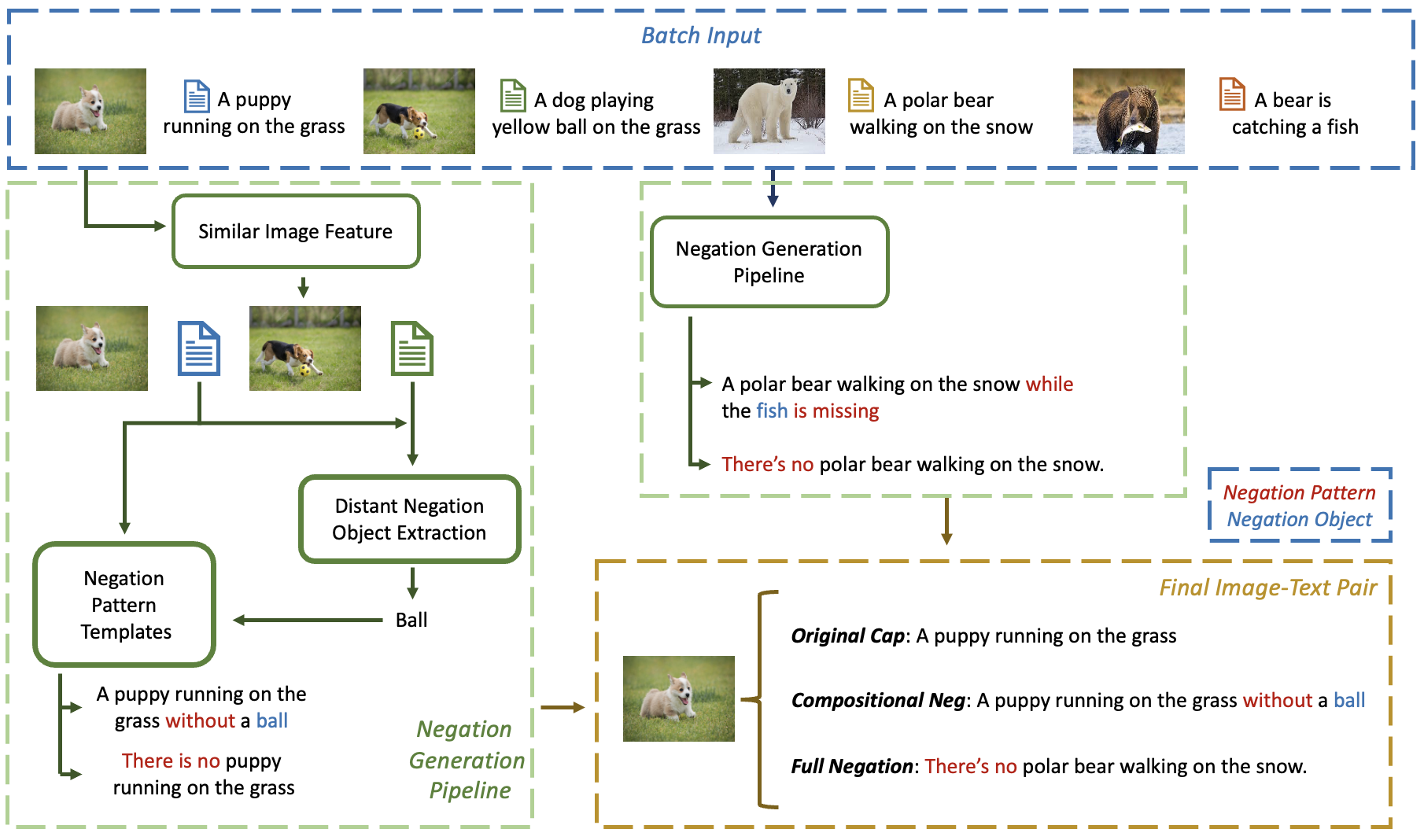}
\caption{\textbf{Training Procedure of \textit{TNG-CLIP}}. The diagram shows the data generation pipeline during the training for one sample in the batch. For an image-text pair, $P_o$, the most similar image pair, $P_s$ is selected by the cosine similarity of their embedded image features. The captions from $P_o$ and $P_s$ are used to find the negation object and generate two types of negation captions. The final image-text set, $S_i$, for $i^{th}$ image-text pair will be composed of one image, $I_i$, one original caption, $T_{o_i}$, one compositional negation caption, $T_{nc_i}$, and one full negation caption, $T_{nf_j}$ from another random sample. }
\label{fig:Pipeline_Diagram}
\end{figure*}

% \subsection{Foundation Model Negation Understanding}
While recent foundation models, including LLMs and VLMs, have achieved remarkable success across diverse downstream tasks, their ability to handle negation semantics remains limited. In the scope of large-scale foundation models, the study of negation understanding starts from language-only setting, where large language models, instead of vision-language models, are focused. \citeauthor{truong2023languagemodelsnaysayersanalysis} shows the LLM's insensitivity of negation by evaluating SOTA LLMs \citep{GPT3,instructgpt,FLAN-T5} on diverse text-only negation benchmarks \citep{hossain-etal-2020-analysis,geiger-etal-2020-neural,truong-etal-2022-another}. \citeauthor{zhang2023positivescalingnegationimpacts} mentions that scaling-up the size of LLM fails to tackle negation tasks. Also, \citeauthor{varshney2024investigatingaddressinghallucinationsllms} analyze and tackle the issue of negation in LLM hallucinations, which also emphasizes the significance of negation understanding in LLMs. 

On the other hand, the negation study in VLMs is mainly focused on CLIP \citep{radford2021learningtransferablevisualmodels}. For example, \citeauthor{quantmeyer2024doesclipprocessnegation} conduct experiments and visualize where and how does CLIP model process negation information in each layers. To make CLIP model understand negation, methods \citep{park2025knownobetterdatadriven,conclip,negbench} adopt LLMs to generate negation caption, based on existing image-text pair datasets, to fine-tune the CLIP for negation understanding. However, generating million-scale negation caption with LLM is extremely time- and compute-consuming, and the negation caption is associated with fixed negation object. For example, when a image is paired with the negation caption \textit{"A dog not with a boy"},  the word \textit{"boy"} can be substituted with plenty of potentially-existing objects such as \textit{"cat"}, \textit{"ball"}, \textit{"food"} and so on. 
\begin{comment}
    More over, existing evaluations of negation fine-tuned CLIP models are primarily limited to negation-aware image-text matching and retrieval benchmarks, with little to no assessment on broader downstream tasks which CLIP can be adopted, such as text-to-image generation, where the CLIP text encoder serves as the semantic input for guiding image generation.
\end{comment}

% \subsection{Fine-tuning with Dynamic Dataset} 
Instead of relying on a fixed and stationary dataset throughout   training, some methods explore the application of dynamic and non-stationary datasets during training process~\citep{wang2019dynamicsentencesamplingefficient,cai2023task,jiang2024adaptive,B_ther_2025,cheng2025dataefficient}, which is an effective strategy to improve model robustness, generalization, and training efficiency. Inspired by the idea of dynamic dataset training, we generate similar but different negation captions for the same image in every epoch of training, which enhances the diversity of the dataset. Thus, models can learn negation semantics via the absence of multiple negation objects to improve robustness and generalization.

\section{Training-Time Negation Data Generation for Negation Understanding}

\label{TNG-CLIP}
To make CLIP learn negation semantics with diverse datasets and without the burden of time- and compute-consuming LLM-based negation caption generation, we present our novel training pipeline, \textbf{T}raining-Time \textbf{N}egation Data \textbf{G}eneration for CLIP (\textit{TNG-CLIP}), such that we generate image-text sets with form
    <$I, T_o, T_{nc}, T_{nf}$>,
from the given image-text pair <$I, T_o$>, where $I$ and $T_o$ represent the provided image and the original (non-negation) caption in the image-text pair dataset, while $T_{nc}$ and $T_{nf}$ represent the two types of generated negation captions: \textbf{compositional negation caption} and \textbf{full negation caption}, discussed in Sec \ref{template}.

\subsection{Training time data augmentation}
We propose a novel negation data-generation pipeline that the negation captions are formed during each batch of training procedure. The negation data generation pipeline for one image in the batch is shown in Figure \ref{fig:Pipeline_Diagram}. Overall, for a given image-text pair, $P_o$, we will first find another similar image-text pair, $P_s$, select the negation object, $O_n$, and generate corresponding negation captions, $T_{nc}$ and $T_{nf}$ with the randomly-chosen negation pattern template and form the image-text set, $S$.

\subsubsection{Find similar image-text pairs}
To form a semantically reasonable compositional negation caption, $T_{nc}$, we need to find a proper negation object, $O_n$, that can be potentially fitted into the original caption, $T_o$. For example, we want $T_{nc}$ to be "\textit{A dog running with no boy around}", instead of "\textit{A dog running with no whale around}", which is semantically unlikely. Previous methods \citep{park2025knownobetterdatadriven,alhamoud2025visionlanguagemodelsunderstandnegation} acquire the proper negation object, $O_n$, through the reasoning of LLM to find the possible object that might appear in the image but is actually absent. For efficiency, we avoid the use of an LLM, and propose to find the possible $O_n$ of the image-text pair, $P_o$, from its most similar images-text pair, $P_s$, in the same batch. Thus, the first step is to find the $P_s$ for every $P_o$ via cosine similarity, between the embedded image features.

Given a visual encoder $E_v(\cdot)$, a batch of images $\textit{I}_b$ is encoded into the corresponding visual features 
\begin{equation}
    \textit{V}_b=E_v(\textit{I}_b), \textit{V}_b\in \mathbb{R}^{B\times D},
\end{equation}
where $B$ is the batch size and $D$ is the hidden dimension of image feature. For $i^{th}$ image feature, $V_{b_i}$, we apply cosine similarity
\begin{equation}
    \textit{V}_{bs_i} = \arg\max_{\textit{V}_j}cos\_sim(\textit{V}_{b_i},\textit{V}_j)
\end{equation}
to find the most similar image feature, $\textit{V}_{bs_i}$, and keep track of the most similar image-text pair, $P_{s_i}$, associated with the image feature $\textit{V}_{bs_{i}}$.

\subsubsection{Select negation object}
After having $P_s$ for each image-text pair, $P_o$, we aim to find the negation object, $O_n$, exists in $P_s$'s caption that does not exist in the caption of $P_o$. For caption in $P_s$, we employ Natural Language Tool Kit~\citep{bird2009nltk} to extract the POS tag of every word, and only keep those represent nouns. To avoid selecting the object which is too semantically close to the words in original caption and cause conflict, we use WordNet \citep{wordnet} and its hand-curated symbolic network to select the negation object, $O_n$, with furthest semantics to those words in the original caption.

\begin{comment}
\begin{algorithm}[]
\caption{Negation Object Selection}
\KwIn{Original Caption $T_o$, Similar Caption $C_S$}

$\text{noun\_list}_o \gets \operatorname{nltk}(\text{O}_c, \texttt{"NN"})$\;
$\text{noun\_list}_s \gets \operatorname{nltk}(\text{O}_s, \texttt{"NN"})$\;
\ForEach{$w_s \in \text{noun\_list}_s$}{
    $d(w_s) \gets 0$\;
    \ForEach{$w_o \in \text{noun\_list}_o$}{
        $d(w_s) \gets d(w_s) + \operatorname{dis}(w_s, w_o)$\;
    }
}
Return $\displaystyle \arg\max_{w_s \in \text{noun\_list}_s} d(w_s)$\;
\label{negation_selection}
\end{algorithm}
\end{comment}

\subsubsection{Template-based negation caption generation}
\label{template}
For every $T_o$ and $O_n$, we employ randomly-chosen negation templates to generate two different types of negation captions: \textbf{compositional negation caption}, $T_{nc}$, and \textbf{full negation caption}, $T_{nf}$. While the compositional negation caption helps model align image with partial negation of a relevant caption, full negation caption makes the image align with the negation semantics of an unrelated caption.  

\begin{enumerate}[leftmargin=*, label=\textbf{\arabic*.}]
    \item \textbf{Compositional Negation Caption:} The negation caption is in the format of 
    "\textit{A} \textit{<negation>} \textit{B}"
    , where \textit{A} denotes the original caption, $T_o$, \textit{B} denotes the negation object, $O_n$, and \textit{<negation>} represents the negation template that combines the two. For example, let \textit{A} denotes "\textit{A dog playing a ball.}", \textit{B} denotes "\textit{Boy}", and \textit{<negation>} denotes "\textit{There is \{caption\}, but not a \{obj\} around.}" The final compositional negation caption, $T_{nc}$, is "\textit{There is a dog playing a ball, but not a boy around.}" To make the generated captions diverse, we use GPT-4o \citep{openai2024gpt4ocard} to generate 46 different negation patterns.
    
    \item \textbf{Full Negation Caption:} The negation caption is in the format of \textit{<negation>} \textit{A}, which is the negation of the entire caption. We use GPT-4o to generate 18 different negation pattern. % eg. "There's no \{caption\} in the image", where we replace the \{caption\} with the current caption.
\end{enumerate}

All the negation patterns and the prompt for GPT-4o to generate them are attached in Appendix \ref{appen-template}.

\subsubsection{Form new image-text set}
Given the original caption, $T_o$, compositional negation caption, $T_{nc}$, and full negation caption, $T_{nf}$, we can now construct the final image-text set, $S$, for training. For each image $I_i$, we associate it with the original caption, $T_{o_i}$, the compositional negation caption, $T_{nc_i}$, and the full negation caption, $T_{nf_j}$, $j\neq i$. Please note that we randomly pick the full negation caption, $T_{nf_j}$, from other image-text pairs, $P_j$. This is because we want to align the negation of the irrelevant captions to the image and contrast the negation of the relevant caption. Finally, the image-text set, $S$, is denoated as
%\begin{equation}
\[
    \text{Image}_i \leftrightarrow \ \left\{
\begin{array}{l}
\text{Original}_i \\
\text{Compositional Negation}_i \\
\text{full negation}_j, j \neq i
\end{array}
\right.
\]
%\end{equation}

\subsection{Asymmetric noise-augmented objective}
\label{sec:objective}
After negation image-text set generation, each image is associated with three captions, which makes the image-text pair imbalanced. Thus, the image-to-text loss, $\mathcal{L}_{i2t}$, and text-to-image loss, $\mathcal{L}_{t2i}$, become asymmetric. We redefine the functionality of both unidirectional loss to serve different purpose. 
\paragraph{Text-to-Image Objective} Given that we have three captions for one image, the similarity matrix will be in shape of $3N\times N$, where $N$ denotes the number of the images. We calculate the $\mathcal{L}_{t2i}$ in a single objective by applying same image alignment to the three captions. The text-to-image objective function is defined as:
%\begin{equation}
\[
    \mathcal{L}_{\text{t2i}} = -\frac{1}{3N} \sum_{j=0}^{3N-1} \log \left( \frac{\exp\left(S_{j, \left\lfloor \frac{j}{3} \right\rfloor} / \tau\right)}{\sum_{i=0}^{N-1} \exp\left(S_{j,i} / \tau\right)} \right),
\]
%\end{equation}
where $S_{j,i}$ denotes the similarity between caption $j$ and image $i$.

\paragraph{Image-to-Text Objective}
 Aligning each image with a negation caption, specifically negation object, is out-of-distribution for pre-trained CLIP because CLIP, which has seen only image–text pairs in which almost all textual components are visually grounded, with no explicit representation of negation. As a result, the pre-trained model struggles to align negation semantics or irrelevant objects with the image. Fine-tuning pre-trained model on such OOD task might lead to worse performance, because fine-tuning can achieve worse accuracy, by overfitting, when
the pretrained models are good and the downstream task distribution shift is large, supported by theory from \citep{kumar2022finetuningdistortpretrainedfeatures}. To solve the above obstacle of overfitting, we introduce label noise to improve the generalization and robustness of the model, inspired by the related works \citep{rolnick2018deeplearningrobustmassive,xie2020selftrainingnoisystudentimproves,chen2025impactnoisysupervisionfoundation}. We modified the image-to-text loss such that the text labels are randomly aligned with the image to introduce noise to the objective function. The $\mathcal{L}_{i2t}$ is:
%\begin{equation}
\[
    \mathcal{L}_{\text{i2t}} = -\frac{1}{N} \sum_{i=0}^{N-1} \log \left( \frac{\exp\left(S_{i, y_i} / \tau\right)}{\sum_{j=0}^{3N - 1} \exp\left(S_{i,j} / \tau\right)} \right),
\]
%\end{equation}
where $y_i \sim \mathcal{U}(\{0, 1, \dots, 3N - 1\})$ is a random selected label across all the captions labels.

\paragraph{Combined Objective}
By introducing noise to $\mathcal{L}_{i2t}$, we only have uni-directional $\mathcal{L}_{t2i}$ helping align negation captions to image. This approach is possible because we freeze the visual encoder during the training, following previous works \citep{conclip,park2025knownobetterdatadriven}. Because the visual encoder is fixed, the visual feature is not updated during image-to-text alignment training, and the model only learn to update text features closer to the pre-trained visual features. The final objective function is then defined as:
\[
    \mathcal{L} = \frac{1}{2}(\mathcal{L}_{i2t} + \mathcal{L}_{t2i}).
\]

The further analysis of the objective function is presented in Appendix \ref{appen-objective}.

\section{Negation Text-to-Image Generation Benchmark}
While negation is an essential part of natural language understanding, a well-designed image generative model should be capable of understanding what to generate and what to avoid. To analyze the generative models' performance on negation prompts, \citeauthor{park2025knownobetterdatadriven} proposed negation-aware image generation experiments with only 107 negation prompts, containing simple naive negation pattern of \textit{"no", "not", "without"}. To enable systematic analysis, we design the first negation-based text-to-image generation benchmark, \textsc{Neg-TtoI}, with examples in Table \ref{table:TtoI-example}. It contains 2000 evaluation samples in the form of <$p$,$q_p$,$q_n$,$a_p$,$a_n$>, where $p$ is the prompt mentioning both desired and undesired objects, $q_p$ is positive question about the existence of desired objects, $q_n$ is the negative question about the absence of undesired objects, and $a_p$ and $a_n$ are the answer to $q_p$ and $q_n$.

%in which each text prompt includes both desired and undesired objects, structured as a compositional sentence. To evaluate the image generation quality, each compositional sentence is associated with one positive question, that questions the existence of desired objects, and one negative question, that questions the absence of undesired objects. Examples can be found in Table \ref{table:TtoI-example}.

\begin{table*}[ht]
\centering
\small
\setlength{\tabcolsep}{6pt}
\renewcommand{\arraystretch}{1.2}
\begin{tabular}{c||c|c|c|c||c|c}
\toprule
\textbf{Model} & \textbf{Avg.} & \textbf{Affirmation} & \textbf{Negation}  & \textbf{Hybrid} & \textbf{R@5}  & \textbf{Neg-R@5} \\
\midrule
CLIP (Pretrained) & 16.28 & 21.89 & 16.89 & 9.99 & 54.76 & 47.92 \\
CoN-CLIP & 15.70 & 0.05 & 36.73 & 11.97 & 51.91 & 48.22 \\
NegCLIP & 10.21 & 9.97 & 19.76 & 1.83 & \textbf{68.73} & \textbf{64.41} \\
CLIP (CC12MNegFull) & 46.9 & 56.49 & 41.71 & 42.29 & 54.20 & 51.90\\
\textbf{TNG-CLIP (Ours)} & \textbf{52.5} & \textbf{68.75} & \textbf{44.75} & \textbf{43.29} & 62.00 & 61.11\\
\bottomrule
\end{tabular}
\caption{Result on Negbench MSCOCO image dataset on image-to-text matching and text-to-image retrieval tasks. \textbf{R@5} refers to the Top-5 accuracy on original (non-negation) MSCOCO-Caption dataset, while \textbf{Neg-R@5} refers to the Top-5 accuracy on negation MSCOCO-Caption dataset from NegBench. }
\label{Negbench-MSCOCO}
\end{table*}

\begin{table*}[h]
\centering
\small
\setlength{\tabcolsep}{6pt}
\renewcommand{\arraystretch}{1.2}
\begin{tabular}{c||c|c|c|c}
\toprule
\textbf{Model} & \textbf{Avg.}  & \textbf{Affirmation}  & \textbf{Negation}  & \textbf{Hybrid} \\
\midrule
CLIP (Pretrained) & 14.47 & 31.96 & 8.34 & 14.97 \\
CoN-CLIP & 22.36 & 0.01 & 27.67 & 24.14 \\
NegCLIP & 8.50 & 22.58 & 8.62 & 4.08 \\
CLIP (CC12MNegFull) & 52.65 & 73.75 & 35.69 & 62.34\\
\textbf{TNG-CLIP (Ours)} & \textbf{59.23} & \textbf{85.92} & \textbf{36.39} & \textbf{72.80} \\
\bottomrule
\end{tabular}
\caption{Result of Negbench image-to-text matching on VOC2007 image dataset}
\label{Negbench-voc2007}
\end{table*}

\subsection{Negation prompts generation pipeline}
We follow the procedure of previous works \citep{park2025knownobetterdatadriven,negbench} to generate prompts and questions via LLM. We use LLM instead of our negation generation pipeline in Sec \ref{TNG-CLIP} because (1) the scale of our evaluation benchmark is much smaller than the scale of training dataset, and (2) we only generate the benchmark prompts and questions once, without the necessity of iterative negation data generation over epochs, which makes the LLM time- and compute-affordable. 

We use the MS-COCO Caption
\citep{chen2015microsoftcococaptionsdata} as the base dataset. The goal of our caption generation pipeline is to transform each caption, which describes the existing scene or objects in the image, into a negation-style caption in which certain elements are explicitly described as absent. To efficiently manipulate the caption with complicated semantics, we leverage GPT-4o \citep{openai2024gpt4ocard} in a multi-step manner from negation prompt generation, evaluation questions generation and quality verification.

\begin{enumerate}[leftmargin=*, label=\textbf{\arabic*.}]

\item \textbf{Negation Prompt Generation:} For every input caption, we ask LLM to identify a random scene or object that is mentioned in the original caption. The selected scene or object will be used as the negation object to generate negation caption. Once we have the original caption and the negation object, we prompt LLM to rewrite the original caption such that the object should be semantically absent from the original caption.

\item \textbf{Evaluation Question Generation} For every negation prompt, we prompt LLM to identify the positive semantics and negative semantics in the sentence while discard the negation pattern. For example, given a negation caption \textit{"A dog playing a yellow ball while there is no man walking around"}, the positive semantics will be \textit{"A dog playing a yellow ball"}, while the negative semantics will be \textit{"man walking around"}. Both the positive semantics and negative semantics are combined with "Is there...?" to form the questions $q_p$ and $q_n$.

\item \textbf{Question Quality Verification}
Although GPT-4o is one of the SOTA LLMs for semantic understanding, it still might generate text that are semantically incorrect. Thus, verification is necessary to prevent the improper generation. Given the negation prompt, $p$, positive question, $q_p$, and negative question $q_n$,  we prompt the LLM to ask whether the semantics in the $q_p$ is stated positively in $p$, and whether the semantics in the $q_n$ is stated negatively in $p$ with the negation semantics. If the LLM's answer for both questions are correct, the negation data sample will be kept, otherwise it will be discarded.

\end{enumerate}
In the end, \textsc{Neg-TtoI} contains 2000 valid samples, selected from 2500 candidates.

\subsection{Evaluation metrics}
Unlike image-text matching or retrieval tasks such that the explicit ground truth can be found, evaluating image generation task is relatively subjective. Inspired by \citep{park2025knownobetterdatadriven,hu2023tifaaccurateinterpretabletexttoimage}, we employ GPT-4o \citep{openai2024gpt4ocard} to evaluate the existence and absence of the objects. Given a image generated using negation caption as prompt and the positive question and negative question, we evaluate the model's generation quality via the metric of \textbf{Compositional Accuracy}: it's \textbf{True} if the LLM answers "yes" on positive question and "no" on negative question at the same time.

\section{Experiments}
To show the capability of our proposed method on multiple downstream tasks, we evaluate our model on negation tasks including image-to-text matching, text-to-image retrieval and text-to-image generation. Our goal is to assess \textit{TNG-CLIP}'s negation semantics understanding via multiple benchmarks and show its generalization and capacity on diverse negation-based scenarios. In the paper, all experiments are performed on a single Nvidia A40 GPU with batch size of 128 and learning rate of 5e-6.

\subsection{Matching \& retrieval evaluation}
To evaluate the negation understanding ability of \textit{TNG-CLIP}, we present the experiments on image-to-text matching and text-to-image retrieval tasks. 

\paragraph{Benchmarks} We employ the following benchmarks to evaluate the model's performance:
\begin{itemize}
    \item Valse-Existence~\citep{Parcalabescu_2022_valse} benchmark evaluates the model's performance on negation imaget-to-text matching task. Given a image and two text description about the presence and absence of an object in the image, \textit{e.g. "There is animal in the image"}/\textit{"There is no animal in the image"}, the model should select the best-matched text.
    
    \item NegBench~\citep{alhamoud2025visionlanguagemodelsunderstandnegation}  benchmark is a comprehensive benchmark to evaluate the negation understanding of models on variant image-to-text matching and text-to-image retrieval tasks. It includes negation-based matching tasks based on both MS-COCO\citep{chen2015microsoftcococaptionsdata} and VOC2007\citep{pascal-voc-2007} datasets, a text-to-image retrieval task based on MS-COCO evaluation dataset, where the captions are converted into compositional negation style. In the matching task, images are paired with four different captions of three categories: \textit{Affirmation} for \textit{"It include A and B."}, \textit{Negation} for \textit{"Does not include A and B."}, and \textit{Hybrid} for \textit{"Include A but not B."}. 
\end{itemize}

\begin{table}[h]
\centering
\begin{tabular}{c|c}
\toprule
 \textbf{Model} & \textbf{Accuracy}   \\
\midrule
CLIP (Pretrained) & 65.16 \\

NegCLIP & 73.22 \\
CoN-CLIP & 74.15\\
CLIP (CC12MNegFull) & 76.21 \\
NegationCLIP & 80.15 \\
\textbf{TNG-CLIP (Ours)} & \textbf{81.64}\\
\bottomrule
\end{tabular}
\caption{Valse-Existence Image-to-Text Matching}
\label{valse_existnce}
\end{table}

\paragraph{Baselines}
To evaluate the performance of our method, we compare it against several existing baseline methods for CLIP's negation understanding, including \textit{pretrained-CLIP}~\citep{radford2021learningtransferablevisualmodels}, \textit{NegCLIP}~\citep{negclip}, \textit{CoN-CLIP}~\citep{conclip}, and CLIP fine-tuned on \textit{CC12M-NegFull}~\citep{negbench}. For fair comparison, all of the methods are initialized based on pre-trained CLIP ViT-B/32 model.

\paragraph{Comparison Experiments}
We present the matching and retrieval task of NegBench-MSCOCO in table \ref{Negbench-MSCOCO} and the matching task of NegBench-VOC2007 in table \ref{Negbench-voc2007}. From the tables, we observe that previous methods are lack of generalization on negation-based tasks, but only focus on the negation understanding of specific tasks. For example, \textit{CoN-CLIP}'s performance on matching (affirmation) task is 0.05 and 0.29 on MSCOCO and VOC2007 datasets, which indicates that the method is biased such that it sacrifices the CLIP's performance on non-negation performance for negation improvement. For \textit{NegCLIP}, even though it get the best score on retrieval task, we observe that the affirmation performance is lower than that of the \textit{pretrained-CLIP}, and its performance on matching (hybrid) is low. On the other hand, the \textit{CC12M-NegFull} fine-tuned CLIP presents general improvement of different tasks, indicating its capability of diverse negation tasks. Our method, \textit{TNG-CLIP}, even though slightly underperforms the \textit{NegCLIP} model on retrieval tasks, achieves SOTA performance on all the matching tasks, shows its generalization and high-performance on diverse scenarios.

Similarly, the evaluation on Valse-Existence dataset, in Table \ref{valse_existnce}, further proofs \textit{TNG-CLIP}'s, capability of negation understanding. While the benchmark is first used by \textit{NegationCLIP}~\citep{park2025knownobetterdatadriven} and achieves promising result of 80.15 on CLIP ViT-B/32 based models, our method gets better performance, 81.64, which is higher than all other negation-understanding CLIP baselines.

\begin{table}[]
\centering
\begin{tabular}{c|c}
\toprule
 \textbf{Strategy} & \textbf{Avg. Acc.} \\
\midrule
dynamic dataset & 51.61 {\small $\pm$ 0.96} \\
fixed dataset & 49.52 {\small $\pm$ 1.27} \\
\bottomrule
\end{tabular}
\caption{Effect of using dynamic dataset. Evaluation on NegBench-MSCOCO image-to-text matching task.}
\label{Dynamic_effect}
\end{table}
\paragraph{Effectiveness of Dynamic Dataset}
The training-time data generation pipeline generates the negation caption based on the other image-text pairs in the same batch, which makes the negation caption of same image different in every epoch. We analyze the effect of such dynamic dataset and compare how the performance differs from using fixed dataset. We store the image-text set, $S$, generated in each training epoch for every epoch as the fixed dataset. We then use the fixed dataset to replace the data generation pipeline to fine-tune the CLIP model. To get statistically significant comparison result, we repeat the \textit{TNG-CLIP}'s training procedure for 10 times and use 10 fixed dataset collected from different training epochs to fine-tune pre-trained CLIP with same objective function and hyper-parameters. We present the mean and standard deviation in Table \ref{Dynamic_effect}. We observe that the performance of \textit{TNG-CLIP} is higher than using fixed dataset, and the standard deviation is also smaller than the fixed one. We explain such phenomenon as the CLIP's fine-tuning on fixed dataset constrains the model's negation understanding to specific \textit{<caption, negation object>} pair, thus harms the generalization of the model on negation tasks, leading to lower mean accuracy. At the same time, the data variance among every epoch for \textit{TNG-CLIP} works as a natural regularization to prevent overfitting and memorizing incorrect correlation, thus lead to smaller standard variance.

More analytic experiments are in Appendix \ref{appen-time} and \ref{appen-image-classificaiton}.

\subsection{Text-to-Image Generation}

\begin{table}[h]
\centering
\scalebox{0.85}{
\begin{tabular}{c||c||c}
\toprule
 \textbf{Model} & \textbf{Arch.} & \textbf{Acc.}  \\
\midrule
SD-1.5 & ViT-L/14 & 32.60 \\
SDXL-1.0 & ViT-L/14  & 27.45 \\
SD-1.5 w/ CoN-CLIP & ViT-L/14 & 28.40\\
SD-1.5 w/ TNG-CLIP (ours) &  ViT-L/14 & \textbf{45.65} \\
\midrule
pretrained-CLIP + proj & ViT-B/32  & 28.25\\
NegCLIP + proj & ViT-B/32 & 33.85 \\
CoN-CLIP + proj & ViT-B/32 & 24.05 \\
CC12MNegFull + proj & ViT-B/32 & 36.95 \\
TNG-CLIP + proj (ours) & ViT-B/32 & \textbf{41.70}\\
\bottomrule

\end{tabular}
}
\caption{Image Generation on \textsc{Neg-TtoI} benchmark}
\label{neg-TtoI}
\end{table}

\subsubsection{CLIP for Image Generation Task}
Although CLIP model is mostly used to do image-text matching tasks, it can be applied to text-to-image generation tasks indirectly. For example, the text encoder from stable diffusion model is the original copy of CLIP ViT-L/14's text encoder \citep{rombach2022highresolutionimagesynthesislatent}. To evaluate the negation understanding of CLIP in text-to-image generation field, \citeauthor{park2025knownobetterdatadriven} provides a simple yet effective way, by replacing the original text encoder from stable diffusion model with their proposed negation-aware CLIP. This direct substitution is possible because they fine-tune only the text encoder, preserving the original image embedding space and maintaining the text feature alignment with it.

\subsubsection{Experiment Setup}
Following the strategy mentioned above, we fine-tuned our \textit{TNG-CLIP} from pretrained CLIP ViT-L/14 model, and replace the original stable diffusion model's text encoder with ours.

However, most baseline methods are fine-tuned only on CLIP ViT-B/32 model, it is difficult to do the direct substitution due to the mismatch of output feature dimension. To tackle such issue, we attach a MLP projector after the frozen text encoder, and perform knowledge distillation between CLIP ViT-L/14's text encoder acts and CLIP ViT-B/32's text encoder with projector, to align the output of projected CLIP ViT-B/32 text encoder similar to that of CLIP ViT-L/14 text encoder. We perform add MLP to all the baseline methods and fine-tune the MLP, with text encoder frozen, on the same dataset, MS-COCO Caption~\citep{chen2015microsoftcococaptionsdata}.

\subsubsection{Experiment Analysis}
The comparison results on \textsc{Neg-TtoI} benchmark are presented in Table \ref{neg-TtoI}. The upper table shows the comparison with CLIP ViT-L/14's text encoder architecture. We choose SD-1.5~\citep{rombach2022highresolutionimagesynthesislatent} as the generative model backbone and replace its text encoder with that of ours and \textit{CoN-CLIP}'s. All the experiment here are the zero-shot performance on \textsc{Neg-TtoI} benchmark. We observe that among the all, using our \textit{TNG-CLIP}'s text encoder achieves the best accuracy, indicating its outstanding capability of handling negation feature for image generation. On the other hand, the accuracy of \textit{CoN-CLIP} is lower than original stable diffusion model, which shows its deficiency on image generation task. 

The lower table presents the accuracy of SD-1.5 by replacing its text encoder with the combination of CLIP ViT-B/32 based architecture and the fine-tuned MLP projector. Noticing that the accuracy of our method using CLIP ViT-B/32's text encoder is 41.70, while that for using CLIP ViT-L/14's text encoder is 45.65, showing that the projected ViT-B/32 text encoder is not as effective as ViT-L/14's text encoder, and is only used for the purpose of providing accessible and fair comparison between the baselines on image generation task. Among the all, our method's text encoder still achieves the best accuracy, and the clip fine-tuned with \textit{CC12MNegFull}~\citep{negbench} is the second best, similar with its performance in image-text matching tasks.

We provide more detailed image generation task analysis in Appendix \ref{appendix-image-generation}.

\begin{comment}
\begin{table*}[h]
\centering
\small
\setlength{\tabcolsep}{6pt}
\renewcommand{\arraystretch}{1.2}
\begin{tabular}{c||c|c|c|c||c|c}
\toprule
\textbf{Model} & \textbf{Avg.} (\uparrow) & \textbf{Affirmation} (\uparrow) & \textbf{Negation} (\uparrow) & \textbf{Hybrid} (\uparrow) & \textbf{R@5} (\uparrow) & \textbf{Neg-R@5} (\uparrow)\\
\midrule
CLIP (Pretrained) & 20.10 & 29.85 & 16.99 & 13.15 & 50.59 & 45.80 \\
ConCLIP & 20.30 & 0.29 & \textbf{35.12} & 0.25 & 48.19 & 40.09 \\
NegCLIP & 15.30 & 16.11 & 22.09 & 6.73 & \textbf{53.70} & \textbf{50.99} \\
CLIP (CC12MNegFull) & 35.60 & 51.34 & 31.16 & 24.03 & 46.90 & 43.90\\
\textbf{Ours} & \textbf{46.40} & \textbf{81.91} & 18.98 & \textbf{40.06} & 48.69 & 49\\
\bottomrule
\end{tabular}
\caption{Result on Negbench MSRVTT video dataset}
\label{Negbench-MSRVTT}
\end{table*}
\end{comment}
\section{Discussion \& Conclusion}
In this paper, we focus on the critical problem of improving negation understanding for CLIP. Instead of using pre-generated fixed negation dataset, we propose a training-time negation data generation pipeline to generate dynamic negation caption during the training time, addressing the time- and compute- inefficiency problem of previous dataset. We also show that using dynamic negation caption during the training can improve mdoel's generalization and boost the performance of negation fine-tuned CLIP. On the other hand, we propose the first negation-aware text-to-image generation evaluation benchmark to expand the horizon of negation-related benchmarks. Overall, our work underscores the negation understanding in the study of vision language model, and call for the wider exploration of negation-aware model in diverse tasks.

\section{Limitations}
In this paper, we propose a negation-aware CLIP, \textit{TNG-CLIP}, trained via the novel efficient training-time negation data generation pipeline. We also propose a negation text-to-image generation benchmark, \textsc{Neg-TtoI}, to evaluate the capability of generative model's performance with negation semantics. However, although we have shown the performance and generalization of \textit{TNG-CLIP} via multiple benchmarks, we see the limit of our paper:

\begin{itemize}
    \item In the paper, we mainly focus on the negation understanding of CLIP model. As the lack of negation understanding is an overall challenge among all vision language models, further exploration on negation-awareness of diverse VLMs is necessary.
    \item The training-time negation data generation pipeline is currently limited to image-text pair dataset, which is adopted to apply contrastive learning. Our negation data generation pipeline has the potential to be extended beyond image-text pairs, eg. visual question answering dataset, thus supports the negation-awareness training with objective function other than contrastive loss.
\end{itemize}

\bibliography{custom}

\begin{thebibliography}{39}
\providecommand{\natexlab}[1]{#1}

\bibitem[{Alhamoud et~al.(2025{\natexlab{a}})Alhamoud, Alshammari, Tian, Li, Torr, Kim, and Ghassemi}]{negbench}
Kumail Alhamoud, Shaden Alshammari, Yonglong Tian, Guohao Li, Philip Torr, Yoon Kim, and Marzyeh Ghassemi. 2025{\natexlab{a}}.
\newblock \href {https://arxiv.org/abs/2501.09425} {Vision-language models do not understand negation}.
\newblock \emph{Preprint}, arXiv:2501.09425.

\bibitem[{Alhamoud et~al.(2025{\natexlab{b}})Alhamoud, Alshammari, Tian, Li, Torr, Kim, and Ghassemi}]{alhamoud2025visionlanguagemodelsunderstandnegation}
Kumail Alhamoud, Shaden Alshammari, Yonglong Tian, Guohao Li, Philip Torr, Yoon Kim, and Marzyeh Ghassemi. 2025{\natexlab{b}}.
\newblock \href {https://arxiv.org/abs/2501.09425} {Vision-language models do not understand negation}.
\newblock \emph{Preprint}, arXiv:2501.09425.

\bibitem[{Bird et~al.(2009)Bird, Klein, and Loper}]{bird2009nltk}
Steven Bird, Ewan Klein, and Edward Loper. 2009.
\newblock \emph{Natural language processing with Python: analyzing text with the natural language toolkit}.
\newblock " O'Reilly Media, Inc.".

\bibitem[{Brown et~al.(2020)Brown, Mann, Ryder, Subbiah, Kaplan, Dhariwal, Neelakantan, Shyam, Sastry, Askell, Agarwal, Herbert-Voss, Krueger, Henighan, Child, Ramesh, Ziegler, Wu, Winter, Hesse, Chen, Sigler, Litwin, Gray, Chess, Clark, Berner, McCandlish, Radford, Sutskever, and Amodei}]{GPT3}
Tom~B. Brown, Benjamin Mann, Nick Ryder, Melanie Subbiah, Jared Kaplan, Prafulla Dhariwal, Arvind Neelakantan, Pranav Shyam, Girish Sastry, Amanda Askell, Sandhini Agarwal, Ariel Herbert-Voss, Gretchen Krueger, Tom Henighan, Rewon Child, Aditya Ramesh, Daniel~M. Ziegler, Jeffrey Wu, Clemens Winter, and 12 others. 2020.
\newblock \href {https://arxiv.org/abs/2005.14165} {Language models are few-shot learners}.
\newblock \emph{Preprint}, arXiv:2005.14165.

\bibitem[{Böther et~al.(2025)Böther, Robroek, Gsteiger, Holzinger, Ma, Tözün, and Klimovic}]{B_ther_2025}
Maximilian Böther, Ties Robroek, Viktor Gsteiger, Robin Holzinger, Xianzhe Ma, Pınar Tözün, and Ana Klimovic. 2025.
\newblock \href {https://doi.org/10.1145/3709705} {Modyn: Data-centric machine learning pipeline orchestration}.
\newblock \emph{Proceedings of the ACM on Management of Data}, 3(1):1–30.

\bibitem[{Cai et~al.(2023)Cai, Thomason, and Rostami}]{cai2023task}
Yuliang Cai, Jesse Thomason, and Mohammad Rostami. 2023.
\newblock Task-attentive transformer architecture for continual learning of vision-and-language tasks using knowledge distillation.
\newblock In \emph{Findings of the Association for Computational Linguistics: EMNLP 2023}, pages 6986--7000.

\bibitem[{Changpinyo et~al.(2021)Changpinyo, Sharma, Ding, and Soricut}]{changpinyo2021conceptual12mpushingwebscale}
Soravit Changpinyo, Piyush Sharma, Nan Ding, and Radu Soricut. 2021.
\newblock \href {https://arxiv.org/abs/2102.08981} {Conceptual 12m: Pushing web-scale image-text pre-training to recognize long-tail visual concepts}.
\newblock \emph{Preprint}, arXiv:2102.08981.

\bibitem[{Chen et~al.(2025)Chen, Wang, Tao, Wei, Xie, Sugiyama, Raj, and Wang}]{chen2025impactnoisysupervisionfoundation}
Hao Chen, Zihan Wang, Ran Tao, Hongxin Wei, Xing Xie, Masashi Sugiyama, Bhiksha Raj, and Jindong Wang. 2025.
\newblock \href {https://arxiv.org/abs/2403.06869} {Impact of noisy supervision in foundation model learning}.
\newblock \emph{Preprint}, arXiv:2403.06869.

\bibitem[{Chen et~al.(2015)Chen, Fang, Lin, Vedantam, Gupta, Dollar, and Zitnick}]{chen2015microsoftcococaptionsdata}
Xinlei Chen, Hao Fang, Tsung-Yi Lin, Ramakrishna Vedantam, Saurabh Gupta, Piotr Dollar, and C.~Lawrence Zitnick. 2015.
\newblock \href {https://arxiv.org/abs/1504.00325} {Microsoft coco captions: Data collection and evaluation server}.
\newblock \emph{Preprint}, arXiv:1504.00325.

\bibitem[{Cheng et~al.(2025)Cheng, Li, and Bian}]{cheng2025dataefficient}
Ziheng Cheng, Zhong Li, and Jiang Bian. 2025.
\newblock \href {https://openreview.net/forum?id=2CflgSMLoK} {Data-efficient training by evolved sampling}.

\bibitem[{Chung et~al.(2022)Chung, Hou, Longpre, Zoph, Tay, Fedus, Li, Wang, Dehghani, Brahma, Webson, Gu, Dai, Suzgun, Chen, Chowdhery, Castro-Ros, Pellat, Robinson, Valter, Narang, Mishra, Yu, Zhao, Huang, Dai, Yu, Petrov, Chi, Dean, Devlin, Roberts, Zhou, Le, and Wei}]{FLAN-T5}
Hyung~Won Chung, Le~Hou, Shayne Longpre, Barret Zoph, Yi~Tay, William Fedus, Yunxuan Li, Xuezhi Wang, Mostafa Dehghani, Siddhartha Brahma, Albert Webson, Shixiang~Shane Gu, Zhuyun Dai, Mirac Suzgun, Xinyun Chen, Aakanksha Chowdhery, Alex Castro-Ros, Marie Pellat, Kevin Robinson, and 16 others. 2022.
\newblock \href {https://arxiv.org/abs/2210.11416} {Scaling instruction-finetuned language models}.
\newblock \emph{Preprint}, arXiv:2210.11416.

\bibitem[{Dumitru et~al.(2013)Dumitru, Goodfellow, Cukierski, and Bengio}]{challenges-in-representation-learning-facial-expression-recognition-challenge}
Dumitru, Ian Goodfellow, Will Cukierski, and Yoshua Bengio. 2013.
\newblock Challenges in representation learning: Facial expression recognition challenge.
\newblock \url{https://kaggle.com/competitions/challenges-in-representation-learning-facial-expression-recognition-challenge}.
\newblock Kaggle.

\bibitem[{Everingham et~al.()Everingham, Van~Gool, Williams, Winn, and Zisserman}]{pascal-voc-2007}
M.~Everingham, L.~Van~Gool, C.~K.~I. Williams, J.~Winn, and A.~Zisserman.
\newblock The {PASCAL} {V}isual {O}bject {C}lasses {C}hallenge 2007 {(VOC2007)} {R}esults.
\newblock http://www.pascal-network.org/challenges/VOC/voc2007/workshop/index.html.

\bibitem[{Geiger et~al.(2020)Geiger, Richardson, and Potts}]{geiger-etal-2020-neural}
Atticus Geiger, Kyle Richardson, and Christopher Potts. 2020.
\newblock \href {https://doi.org/10.18653/v1/2020.blackboxnlp-1.16} {Neural natural language inference models partially embed theories of lexical entailment and negation}.
\newblock In \emph{Proceedings of the Third BlackboxNLP Workshop on Analyzing and Interpreting Neural Networks for NLP}, pages 163--173, Online. Association for Computational Linguistics.

\bibitem[{Hodosh et~al.(2013)Hodosh, Young, and Hockenmaier}]{flickr8k}
Micah Hodosh, Peter Young, and Julia Hockenmaier. 2013.
\newblock Framing image description as a ranking task: data, models and evaluation metrics.
\newblock \emph{J. Artif. Int. Res.}, 47(1):853–899.

\bibitem[{Hossain et~al.(2020)Hossain, Kovatchev, Dutta, Kao, Wei, and Blanco}]{hossain-etal-2020-analysis}
Md~Mosharaf Hossain, Venelin Kovatchev, Pranoy Dutta, Tiffany Kao, Elizabeth Wei, and Eduardo Blanco. 2020.
\newblock \href {https://doi.org/10.18653/v1/2020.emnlp-main.732} {An analysis of natural language inference benchmarks through the lens of negation}.
\newblock In \emph{Proceedings of the 2020 Conference on Empirical Methods in Natural Language Processing (EMNLP)}, pages 9106--9118, Online. Association for Computational Linguistics.

\bibitem[{Hu et~al.(2023)Hu, Liu, Kasai, Wang, Ostendorf, Krishna, and Smith}]{hu2023tifaaccurateinterpretabletexttoimage}
Yushi Hu, Benlin Liu, Jungo Kasai, Yizhong Wang, Mari Ostendorf, Ranjay Krishna, and Noah~A Smith. 2023.
\newblock \href {https://arxiv.org/abs/2303.11897} {Tifa: Accurate and interpretable text-to-image faithfulness evaluation with question answering}.
\newblock \emph{Preprint}, arXiv:2303.11897.

\bibitem[{Jiang et~al.(2024)Jiang, Zhou, Feng, Malladi, and Kolter}]{jiang2024adaptive}
Yiding Jiang, Allan Zhou, Zhili Feng, Sadhika Malladi, and J~Zico Kolter. 2024.
\newblock Adaptive data optimization: Dynamic sample selection with scaling laws.
\newblock \emph{arXiv preprint arXiv:2410.11820}.

\bibitem[{Krizhevsky(2009)}]{cifar}
Alex Krizhevsky. 2009.
\newblock \href {https://api.semanticscholar.org/CorpusID:18268744} {Learning multiple layers of features from tiny images}.

\bibitem[{Kumar et~al.(2022)Kumar, Raghunathan, Jones, Ma, and Liang}]{kumar2022finetuningdistortpretrainedfeatures}
Ananya Kumar, Aditi Raghunathan, Robbie Jones, Tengyu Ma, and Percy Liang. 2022.
\newblock \href {https://arxiv.org/abs/2202.10054} {Fine-tuning can distort pretrained features and underperform out-of-distribution}.
\newblock \emph{Preprint}, arXiv:2202.10054.

\bibitem[{Miller(1995)}]{wordnet}
George~A. Miller. 1995.
\newblock \href {https://doi.org/10.1145/219717.219748} {Wordnet: a lexical database for english}.
\newblock \emph{Commun. ACM}, 38(11):39–41.

\bibitem[{OpenAI et~al.(2024)OpenAI, :, Hurst, Lerer, Goucher, Perelman, Ramesh, Clark, Ostrow, Welihinda, Hayes, Radford, Mądry, Baker-Whitcomb, Beutel, Borzunov, Carney, Chow, Kirillov, Nichol, Paino, Renzin, Passos, Kirillov, Christakis, Conneau, Kamali, Jabri, Moyer, Tam, Crookes, Tootoochian, Tootoonchian, Kumar, Vallone, Karpathy, Braunstein, Cann, Codispoti, Galu, Kondrich, Tulloch, Mishchenko, Baek, Jiang, Pelisse, Woodford, Gosalia, Dhar, Pantuliano, Nayak, Oliver, Zoph, Ghorbani, Leimberger, Rossen, Sokolowsky, Wang, Zweig, Hoover, Samic, McGrew, Spero, Giertler, Cheng, Lightcap, Walkin, Quinn, Guarraci, Hsu, Kellogg, Eastman, Lugaresi, Wainwright, Bassin, Hudson, Chu, Nelson, Li, Shern, Conger, Barette, Voss, Ding, Lu, Zhang, Beaumont, Hallacy, Koch, Gibson, Kim, Choi, McLeavey, Hesse, Fischer, Winter, Czarnecki, Jarvis, Wei, Koumouzelis, Sherburn, Kappler, Levin, Levy, Carr, Farhi, Mely, Robinson, Sasaki, Jin, Valladares, Tsipras, Li, Nguyen, Findlay, Oiwoh, Wong, Asdar, Proehl, Yang, Antonow,
  Kramer, Peterson, Sigler, Wallace, Brevdo, Mays, Khorasani, Such, Raso, Zhang, von Lohmann, Sulit, Goh, Oden, Salmon, Starace, Brockman, Salman, Bao, Hu, Wong, Wang, Schmidt, Whitney, Jun, Kirchner, de~Oliveira~Pinto, Ren, Chang, Chung, Kivlichan, O'Connell, O'Connell, Osband, Silber, Sohl, Okuyucu, Lan, Kostrikov, Sutskever, Kanitscheider, Gulrajani, Coxon, Menick, Pachocki, Aung, Betker, Crooks, Lennon, Kiros, Leike, Park, Kwon, Phang, Teplitz, Wei, Wolfe, Chen, Harris, Varavva, Lee, Shieh, Lin, Yu, Weng, Tang, Yu, Jang, Candela, Beutler, Landers, Parish, Heidecke, Schulman, Lachman, McKay, Uesato, Ward, Kim, Huizinga, Sitkin, Kraaijeveld, Gross, Kaplan, Snyder, Achiam, Jiao, Lee, Zhuang, Harriman, Fricke, Hayashi, Singhal, Shi, Karthik, Wood, Rimbach, Hsu, Nguyen, Gu-Lemberg, Button, Liu, Howe, Muthukumar, Luther, Ahmad, Kai, Itow, Workman, Pathak, Chen, Jing, Guy, Fedus, Zhou, Mamitsuka, Weng, McCallum, Held, Ouyang, Feuvrier, Zhang, Kondraciuk, Kaiser, Hewitt, Metz, Doshi, Aflak, Simens, Boyd,
  Thompson, Dukhan, Chen, Gray, Hudnall, Zhang, Aljubeh, Litwin, Zeng, Johnson, Shetty, Gupta, Shah, Yatbaz, Yang, Zhong, Glaese, Chen, Janner, Lampe, Petrov, Wu, Wang, Fradin, Pokrass, Castro, de~Castro, Pavlov, Brundage, Wang, Khan, Murati, Bavarian, Lin, Yesildal, Soto, Gimelshein, Cone, Staudacher, Summers, LaFontaine, Chowdhury, Ryder, Stathas, Turley, Tezak, Felix, Kudige, Keskar, Deutsch, Bundick, Puckett, Nachum, Okelola, Boiko, Murk, Jaffe, Watkins, Godement, Campbell-Moore, Chao, McMillan, Belov, Su, Bak, Bakkum, Deng, Dolan, Hoeschele, Welinder, Tillet, Pronin, Tillet, Dhariwal, Yuan, Dias, Lim, Arora, Troll, Lin, Lopes, Puri, Miyara, Leike, Gaubert, Zamani, Wang, Donnelly, Honsby, Smith, Sahai, Ramchandani, Huet, Carmichael, Zellers, Chen, Chen, Nigmatullin, Cheu, Jain, Altman, Schoenholz, Toizer, Miserendino, Agarwal, Culver, Ethersmith, Gray, Grove, Metzger, Hermani, Jain, Zhao, Wu, Jomoto, Wu, Shuaiqi, Xia, Phene, Papay, Narayanan, Coffey, Lee, Hall, Balaji, Broda, Stramer, Xu, Gogineni,
  Christianson, Sanders, Patwardhan, Cunninghman, Degry, Dimson, Raoux, Shadwell, Zheng, Underwood, Markov, Sherbakov, Rubin, Stasi, Kaftan, Heywood, Peterson, Walters, Eloundou, Qi, Moeller, Monaco, Kuo, Fomenko, Chang, Zheng, Zhou, Manassra, Sheu, Zaremba, Patil, Qian, Kim, Cheng, Zhang, He, Zhang, Jin, Dai, and Malkov}]{openai2024gpt4ocard}
OpenAI, :, Aaron Hurst, Adam Lerer, Adam~P. Goucher, Adam Perelman, Aditya Ramesh, Aidan Clark, AJ~Ostrow, Akila Welihinda, Alan Hayes, Alec Radford, Aleksander Mądry, Alex Baker-Whitcomb, Alex Beutel, Alex Borzunov, Alex Carney, Alex Chow, Alex Kirillov, and 401 others. 2024.
\newblock \href {https://arxiv.org/abs/2410.21276} {Gpt-4o system card}.
\newblock \emph{Preprint}, arXiv:2410.21276.

\bibitem[{Ouyang et~al.(2022)Ouyang, Wu, Jiang, Almeida, Wainwright, Mishkin, Zhang, Agarwal, Slama, Ray, Schulman, Hilton, Kelton, Miller, Simens, Askell, Welinder, Christiano, Leike, and Lowe}]{instructgpt}
Long Ouyang, Jeff Wu, Xu~Jiang, Diogo Almeida, Carroll~L. Wainwright, Pamela Mishkin, Chong Zhang, Sandhini Agarwal, Katarina Slama, Alex Ray, John Schulman, Jacob Hilton, Fraser Kelton, Luke Miller, Maddie Simens, Amanda Askell, Peter Welinder, Paul Christiano, Jan Leike, and Ryan Lowe. 2022.
\newblock \href {https://arxiv.org/abs/2203.02155} {Training language models to follow instructions with human feedback}.
\newblock \emph{Preprint}, arXiv:2203.02155.

\bibitem[{Parcalabescu et~al.(2022)Parcalabescu, Cafagna, Muradjan, Frank, Calixto, and Gatt}]{Parcalabescu_2022_valse}
Letitia Parcalabescu, Michele Cafagna, Lilitta Muradjan, Anette Frank, Iacer Calixto, and Albert Gatt. 2022.
\newblock \href {https://doi.org/10.18653/v1/2022.acl-long.567} {Valse: A task-independent benchmark for vision and language models centered on linguistic phenomena}.
\newblock In \emph{Proceedings of the 60th Annual Meeting of the Association for Computational Linguistics (Volume 1: Long Papers)}, page 8253–8280. Association for Computational Linguistics.

\bibitem[{Park et~al.(2025)Park, Lee, Song, Yu, Jung, and Yoon}]{park2025knownobetterdatadriven}
Junsung Park, Jungbeom Lee, Jongyoon Song, Sangwon Yu, Dahuin Jung, and Sungroh Yoon. 2025.
\newblock \href {https://arxiv.org/abs/2501.10913} {Know "no'' better: A data-driven approach for enhancing negation awareness in clip}.
\newblock \emph{Preprint}, arXiv:2501.10913.

\bibitem[{Plummer et~al.(2016)Plummer, Wang, Cervantes, Caicedo, Hockenmaier, and Lazebnik}]{flickr30k}
Bryan~A. Plummer, Liwei Wang, Chris~M. Cervantes, Juan~C. Caicedo, Julia Hockenmaier, and Svetlana Lazebnik. 2016.
\newblock \href {https://arxiv.org/abs/1505.04870} {Flickr30k entities: Collecting region-to-phrase correspondences for richer image-to-sentence models}.
\newblock \emph{Preprint}, arXiv:1505.04870.

\bibitem[{Quantmeyer et~al.(2024)Quantmeyer, Mosteiro, and Gatt}]{quantmeyer2024doesclipprocessnegation}
Vincent Quantmeyer, Pablo Mosteiro, and Albert Gatt. 2024.
\newblock \href {https://arxiv.org/abs/2407.10488} {How and where does clip process negation?}
\newblock \emph{Preprint}, arXiv:2407.10488.

\bibitem[{Radford et~al.(2021)Radford, Kim, Hallacy, Ramesh, Goh, Agarwal, Sastry, Askell, Mishkin, Clark, Krueger, and Sutskever}]{radford2021learningtransferablevisualmodels}
Alec Radford, Jong~Wook Kim, Chris Hallacy, Aditya Ramesh, Gabriel Goh, Sandhini Agarwal, Girish Sastry, Amanda Askell, Pamela Mishkin, Jack Clark, Gretchen Krueger, and Ilya Sutskever. 2021.
\newblock \href {https://arxiv.org/abs/2103.00020} {Learning transferable visual models from natural language supervision}.
\newblock \emph{Preprint}, arXiv:2103.00020.

\bibitem[{Rolnick et~al.(2018)Rolnick, Veit, Belongie, and Shavit}]{rolnick2018deeplearningrobustmassive}
David Rolnick, Andreas Veit, Serge Belongie, and Nir Shavit. 2018.
\newblock \href {https://arxiv.org/abs/1705.10694} {Deep learning is robust to massive label noise}.
\newblock \emph{Preprint}, arXiv:1705.10694.

\bibitem[{Rombach et~al.(2022)Rombach, Blattmann, Lorenz, Esser, and Ommer}]{rombach2022highresolutionimagesynthesislatent}
Robin Rombach, Andreas Blattmann, Dominik Lorenz, Patrick Esser, and Björn Ommer. 2022.
\newblock \href {https://arxiv.org/abs/2112.10752} {High-resolution image synthesis with latent diffusion models}.
\newblock \emph{Preprint}, arXiv:2112.10752.

\bibitem[{Singh et~al.(2024)Singh, Shrivastava, Vatsa, Singh, and Bharati}]{conclip}
Jaisidh Singh, Ishaan Shrivastava, Mayank Vatsa, Richa Singh, and Aparna Bharati. 2024.
\newblock \href {https://arxiv.org/abs/2403.20312} {Learn "no" to say "yes" better: Improving vision-language models via negations}.
\newblock \emph{Preprint}, arXiv:2403.20312.

\bibitem[{Truong et~al.(2023)Truong, Baldwin, Verspoor, and Cohn}]{truong2023languagemodelsnaysayersanalysis}
Thinh~Hung Truong, Timothy Baldwin, Karin Verspoor, and Trevor Cohn. 2023.
\newblock \href {https://arxiv.org/abs/2306.08189} {Language models are not naysayers: An analysis of language models on negation benchmarks}.
\newblock \emph{Preprint}, arXiv:2306.08189.

\bibitem[{Truong et~al.(2022)Truong, Otmakhova, Baldwin, Cohn, Lau, and Verspoor}]{truong-etal-2022-another}
Thinh~Hung Truong, Yulia Otmakhova, Timothy Baldwin, Trevor Cohn, Jey~Han Lau, and Karin Verspoor. 2022.
\newblock \href {https://doi.org/10.18653/v1/2022.aacl-main.65} {Not another negation benchmark: The {N}a{N}-{NLI} test suite for sub-clausal negation}.
\newblock In \emph{Proceedings of the 2nd Conference of the Asia-Pacific Chapter of the Association for Computational Linguistics and the 12th International Joint Conference on Natural Language Processing (Volume 1: Long Papers)}, pages 883--894, Online only. Association for Computational Linguistics.

\bibitem[{Varshney et~al.(2024)Varshney, Raj, Mishra, Chatterjee, Sarkar, Saeidi, and Baral}]{varshney2024investigatingaddressinghallucinationsllms}
Neeraj Varshney, Satyam Raj, Venkatesh Mishra, Agneet Chatterjee, Ritika Sarkar, Amir Saeidi, and Chitta Baral. 2024.
\newblock \href {https://arxiv.org/abs/2406.05494} {Investigating and addressing hallucinations of llms in tasks involving negation}.
\newblock \emph{Preprint}, arXiv:2406.05494.

\bibitem[{Wang et~al.(2019)Wang, Utiyama, and Sumita}]{wang2019dynamicsentencesamplingefficient}
Rui Wang, Masao Utiyama, and Eiichiro Sumita. 2019.
\newblock \href {https://arxiv.org/abs/1805.00178} {Dynamic sentence sampling for efficient training of neural machine translation}.
\newblock \emph{Preprint}, arXiv:1805.00178.

\bibitem[{Xiao et~al.(2010)Xiao, Hays, Ehinger, Oliva, and Torralba}]{SUN397}
Jianxiong Xiao, James Hays, Krista~A. Ehinger, Aude Oliva, and Antonio Torralba. 2010.
\newblock \href {https://doi.org/10.1109/CVPR.2010.5539970} {Sun database: Large-scale scene recognition from abbey to zoo}.
\newblock In \emph{2010 IEEE Computer Society Conference on Computer Vision and Pattern Recognition}, pages 3485--3492.

\bibitem[{Xie et~al.(2020)Xie, Luong, Hovy, and Le}]{xie2020selftrainingnoisystudentimproves}
Qizhe Xie, Minh-Thang Luong, Eduard Hovy, and Quoc~V. Le. 2020.
\newblock \href {https://arxiv.org/abs/1911.04252} {Self-training with noisy student improves imagenet classification}.
\newblock \emph{Preprint}, arXiv:1911.04252.

\bibitem[{Yuksekgonul et~al.(2023)Yuksekgonul, Bianchi, Kalluri, Jurafsky, and Zou}]{negclip}
Mert Yuksekgonul, Federico Bianchi, Pratyusha Kalluri, Dan Jurafsky, and James Zou. 2023.
\newblock \href {https://arxiv.org/abs/2210.01936} {When and why vision-language models behave like bags-of-words, and what to do about it?}
\newblock \emph{Preprint}, arXiv:2210.01936.

\bibitem[{Zhang et~al.(2023)Zhang, Yasunaga, Zhou, HaoChen, Zou, Liang, and Yeung}]{zhang2023positivescalingnegationimpacts}
Yuhui Zhang, Michihiro Yasunaga, Zhengping Zhou, Jeff~Z. HaoChen, James Zou, Percy Liang, and Serena Yeung. 2023.
\newblock \href {https://arxiv.org/abs/2305.17311} {Beyond positive scaling: How negation impacts scaling trends of language models}.
\newblock \emph{Preprint}, arXiv:2305.17311.

\end{thebibliography}

\appendix

\section{Appendix}
\label{sec:appendix}

\begin{table*}[h]
\centering
\begin{tabular}{c||c|c|c|c|c}
\toprule
 \textbf{Model} & \textbf{Avg.}  & \textbf{Affirmation}  &\textbf{Negation}  & \textbf{Hybrid}  & \textbf{Neg-R@5} \\
\midrule
TNG-CLIP  & 52.50 & 68.75 & 44.75 & 43.29 & 61.11\\
\midrule 
\midrule
\multicolumn{6}{c}{\small\textit{Ablation of Caption Category}} \\
\midrule
w/o compositional & 48.15 & 65.45 & 38.02 & 40.10 & 56.66 \\
w/o full & 51.31 & 75.63 & 24.75 & 51.44 
 & 60.79 \\
w/o original & 46.66 & 60.18 & 45.82 & 37.79 & 59.91 \\
\midrule
\midrule
\multicolumn{6}{c}{\small\textit{Ablation of Noise}} \\
\midrule
$\mathcal{L}_{i2t}$: original & 52.49 & 81.93 & 16.09 & 56.66 & 45.32  \\
$\mathcal{L}_{i2t}$: compositional & 50.13 & 78.05 & 11.97 & 57.40 & 45.39\\
$\mathcal{L}_{i2t}$: full & 40.29 & 45.12 & 44.11 & 31.85 & 48.58\\
$\mathcal{L}_{i2t}$: random of three & 47.92 & 58.66 & 43.26 & 41.10 & 50.66 \\
w/o $\mathcal{L}_{i2t}$ & 46.24 & 59.54 & 36.04 & 42.29 & 50.49 \\ 
\midrule
independent losses & 46.49 & 55.41 & 43.74 & 40.05 & 50.19 \\
\bottomrule

\end{tabular}
\caption{Ablation Study on NegBench MSCOCO matching task}
\label{ablation_loss}
\end{table*}

\subsection{Ablation Study of Asymmetric Noise-Augmented Objective}
\label{appen-objective}
In order to train the negation-aware CLIP for diverse tasks, we propose a novel asymmetric noise-augmented loss that different from the original contrastive loss of CLIP. We exam the contribution of each component in this novel objective function with analytic ablation study. Within the objective function, we split its component to four parts based on the functionality of each:
\begin{itemize}
    \item \textbf{compositional alignment} refers to align the compositional negation caption to the image in $\mathcal{L}_{t2i}$.
    \item \textbf{full alignment} refers to align the full negation caption to the image in the $\mathcal{L}_{t2i}$.
    \item \textbf{original alignment} refers to align the original caption to the image in $\mathcal{L}_{t2i}$.
    \item \textbf{noise alignment} refers to align the random-chose caption to the image in $\mathcal{L}_{i2t}$.
\end{itemize}
The analytic results are presented in Table \ref{ablation_loss}, with the evaluation on Negbench-MSCOCO matching and Negbench-MSCOCO Retrieval tasks. From the table, we observe that by eliminating compositional negation, full negation and original caption from $\mathcal{L}_{t2i}$ separately, the corresponding performance in matching task drops. For example, without original caption, the affirmation accuracy drops from 68.75 to 60.18. At the same time, the accuracy of negation retrieval tasks remains similar, indicating that the components in $\mathcal{L}_{t2i}$ are not the primary factors for it.

We then analyze the effect of random noise in $\mathcal{L}_{i2t}$. Instead of letting image random choose caption, we match the image to its corresponding original, compositional negation and full negation captions as three independent experiments. Additionally, we let images to randomly match one of its corresponding original, compositional negation and full negation caption, and even directly delete the $\mathcal{L}_{i2t}$ loss. Through the experiments, we found that \textbf{without the random noise, performance of retrieval task drops significantly.} This matches the hypothesis we proposed in Sec \ref{sec:objective} that negation dataset is an OOD task for pre-trained CLIP, the direct fine-tuning may cause worse performance.

Lastly, we propose and examine another loss objective: can we split the generated image-text set, $S$, to form three image-text pairs for each of original, compositional negation and full negation, and apply normal contrastive loss on the three independently? We implement such objective function and present it at the bottom of the table. We observe that by doing so, the performance of both matching task and retrieval task are sub-optimal. While there is also no noise label added to the objective training, the worse result on using independent losses, again, emphasizes the importance of adding noise when fine-tuning CLIP on negation-related dataset.

\subsection{More Analysis of Image Generation Experiment}
\label{appendix-image-generation}
\begin{table*}[h]
\centering
\begin{tabular}{c||c||c|c}
\toprule
 \textbf{Model} & \textbf{Arch.} & \textbf{Positive} &\textbf{Negative}  \\
\midrule
SD-1.5 & ViT-L/14 & 80.85 & 41.95  \\
SDXL-1.0 & ViT-L/14 & 87.05 & 32.30 \\
SD-1.5 w/ CoN-CLIP & ViT-L/14 & 46.25 & 72.50\\
SD-1.5 w/ TNG-CLIP (ours) &  ViT-L/14 & 75.80 & 63.05 \\
\midrule
pretrained CLIP + proj & ViT-B/32 & 45.65 & 67.25 \\
NegCLIP + proj & ViT-B/32 & 67.80 & 52.10 \\
CoN-CLIP + proj & ViT-B/32 & 39.45 & 72.65  \\
CC12MNegFull + proj & ViT-B/32 & 53.76 & 71.80  \\
TNG-CLIP + proj & ViT-B/32 & 63.65 & 68.50 \\
\bottomrule

\end{tabular}
\caption{Image Generation on Neg-TtoI benchmark}
\label{neg-TtoI-continue}
\end{table*}

In the image generation task, we observe the inefficiency of original Stable Diffusion and CoN-CLIP in the \textsc{NegTtoI} benchmark. But why this happens? To further explore that, we evaluate the performance of models with two analytic metrics: \textbf{Positive Accuracy} and \textbf{Negative Accuracy}. Given a prompt "generate A without B", \textbf{Positive Accuracy} measures if the image contains A, and \textbf{Negative Accuracy} measures if the image doesn't contain B. The result is presented in Table \ref{neg-TtoI-continue}. In the table, we can observe that for original Stable Diffusion model, the positive accuracy is higher than that of using our method or CoN-CLIP as text encoder, but the negative accuracy is much lower. This explicitly shows that the original text encoder cannot process negation semantics to help avoid the generation of unwanted objects. On the other hand, adopting CoN-CLIP as text encoder can significantly boost the negative accuracy, but at the same time, its performance on positive accuracy becomes low. This indicates the CoN-CLIP model is a biased model towards negation-understanding, while ignores the generalization on other non-negation tasks. 

\subsection{Non-Negation Generalization on Image Classification}
\label{appen-image-classificaiton}
\begin{figure}
\centering
\includegraphics[scale=0.5]{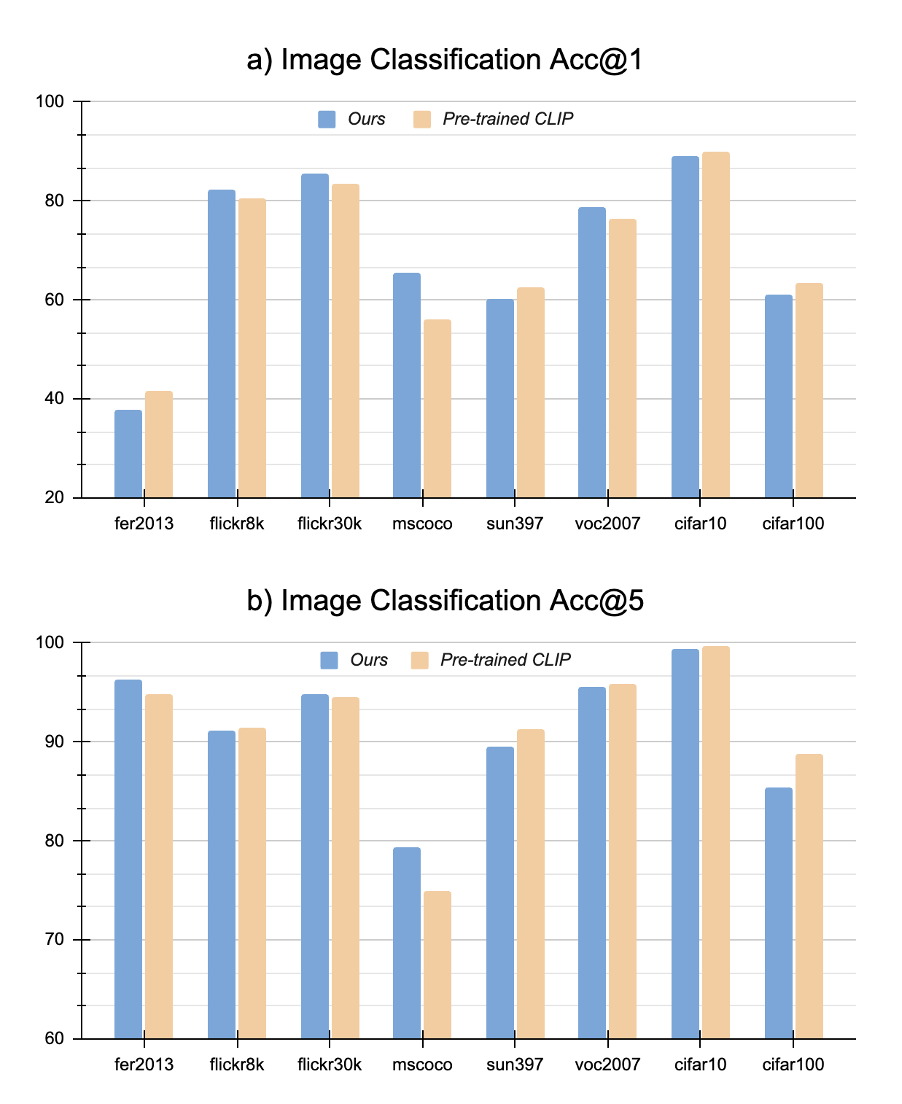}
\caption{The zero shot image classification accuracy of pre-trained CLIP and TNG-CLIP on eight image classification benchmarks. }
\label{fig:Image_classification}
\end{figure}

Although TNG-CLIP is specifically designed for negation understanding, it is important to ensure that its performance on non-negation tasks remains intact, in another word, it should not suffer from catastrophic forgetting on tasks that the original pre-trained CLIP model was capable of handling. Inspired by the experiments from \citep{conclip}, we conduct the zero shot image classification on TNG-CLIP and pre-trained CLIP with eight diverse benchmarks: \textbf{FER2013} \citep{challenges-in-representation-learning-facial-expression-recognition-challenge}, \textbf{Flickr-8K} \citep{flickr8k}, \textbf{Flickr-30K} \citep{flickr30k}, \textbf{MS-COCO} \citep{chen2015microsoftcococaptionsdata}, \textbf{SUN397} \citep{SUN397}, \textbf{VOC2007} \citep{pascal-voc-2007}, \textbf{CIFAR-10} \citep{cifar}, \textbf{CIFAR-100} \citep{cifar}. The top1 and top5 accuracy score is presented in Figure \ref{fig:Image_classification}. In the figure, we observe that the zero-shot performance of TNG-CLIP remains similar with that of pre-trained CLIP, indicating there is no catastrophic forgetting or overfitting to our proposed method. Surprisingly, we also observe that in some cases, such as Flickr-8K, Flickr-30K, MS-COCO and VOC2007 benchmarks, the TNG-CLIP outperforms the pre-trained CLIP, illustrating improving the negation understanding can improve model's performance on general tasks.

\subsection{Time-Efficiency Test}
\label{appen-time}
As we generate data samples during the training stage, does the generation pipeline significantly slower the training process and becomes time-consuming? We compares the average training time per batch on the same GPU device, Nvidia-A40, with and without the data generation pipeline in Table \ref{time_efficiency}. For every batch, the data generation pipeline takes 0.13 sec, which is only 2.55\% slower than without using the data generation pipeline. Thus, adding the data generation pipeline to the training is still time-efficient.
\begin{table}[h]
\centering
\begin{tabular}{c|c}
\toprule
 \textbf{Strategy} & \textbf{Time (sec)} \\
\midrule
w/o data generation & 4.97 \\
w/ data generation & 5.10 \\
data generation & 0.13 \\
\bottomrule
\end{tabular}
\caption{Time Efficiency for Data Generaiton}
\label{time_efficiency}
\end{table}

\subsection{Template-based Negation Pattern}
\label{appen-template}
During the negation caption generation, we use pre-defined LLM-generated negation pattern template to convert the original caption and negation object to compositional negation caption and full negation caption. We present the template we used here in Table \ref{table:complete-template} and Table \ref{table:compositional-template}.

\begin{table*}[h]
\centering
\begin{tabular}{|p{7cm}|p{7cm}|}
\hline
There's no \{cap\} in the image. & No \{cap\} is included in the image. \\
\hline
There is not \{cap\} in the image. & The image does not have \{cap\}. \\
\hline
No \{cap\} is present in the image. & \{cap\} is not present in the image. \\
\hline
\{cap\} is absent. & No \{cap\} is present. \\
\hline
There isn't any \{cap\}. & Not a single \{cap\} can be seen. \\
\hline
The image is without \{cap\}. & The image is lacking \{cap\}. \\
\hline
There appears to be no \{cap\} in the image. & The image does not contain \{cap\}. \\
\hline
There does not exist \{cap\} in the image. & There is nothing about \{cap\}. \\
\hline
There isn't any \{cap\}. & No \{cap\} is seen in the image. \\
\hline
\end{tabular}
\caption{Templates for full negation caption generation, we replace the ${cap}$ with the provided original caption.}
\label{table:complete-template}
\end{table*}

\begin{table*}[h]
\centering
\begin{tabular}{|p{7cm}|p{7cm}|}
\hline
\{cap\} with no \{obj\}. & \{cap\} without \{obj\} \\
\hline
\{cap\} that do not have \{obj\}. & \{cap\} having no \{obj\}. \\
\hline
\{cap\} not include \{obj\}. & \{cap\} excluding \{obj\}. \\
\hline
\{cap\}, but no \{obj\} are present. & \{cap\}, though no \{obj\} can be seen. \\
\hline
\{cap\} without any \{obj\} in sight. & \{cap\} yet no \{obj\} are nearby. \\
\hline
\{cap\} but no \{obj\} are visible. & \{cap\} and no \{obj\} are anywhere around. \\
\hline
\{cap\}, without any \{obj\} in the vicinity. & \{cap\}, with no \{obj\} in the surroundings. \\
\hline
\{cap\}, but no \{obj\} are in the area. & \{cap\}, and no \{obj\} can be found nearby. \\
\hline
\{cap\} in the absence of \{obj\}. & \{cap\}, where no \{obj\} are present. \\
\hline
\{cap\} with an absence of \{obj\}. & \{cap\}, as no \{obj\} are around. \\
\hline
\{cap\}, while lacking any \{obj\}. & \{cap\} but no \{obj\} are engaging. \\
\hline
\{cap\} with no \{obj\} participating. & \{cap\} yet no \{obj\} are interacting. \\
\hline
\{cap\}, as no \{obj\} are involved. & \{cap\}, while \{obj\} remain absent from the scene. \\
\hline
\{cap\} though no \{obj\} can be spotted. & \{cap\} where no \{obj\} are noticeable. \\
\hline
\{cap\} but no \{obj\} are detectable. & \{cap\}, as no \{obj\} are apparent. \\
\hline
\{cap\}, with no sight of any \{obj\}. & No \{obj\} is visible, but \{cap\}. \\
\hline
No \{obj\} can be seen, while \{cap\} happens. & No \{obj\} is present, yet \{cap\} continues. \\
\hline
No \{obj\} appears in sight, but \{cap\} unfolds. & Not a single \{obj\} is noticeable, but \{cap\}. \\
\hline
No trace of \{obj\} can be found, while \{cap\} occurs. & No sign of \{obj\} is apparent, but \{cap\} is happening. \\
\hline
There is no \{obj\} in view, but \{cap\} takes place. & None of the \{obj\} are around, yet \{cap\} continues. \\
\hline
Not even one \{obj\} is nearby, but \{cap\} is ongoing. & No \{obj\} exists in the scene, while \{cap\} happens. \\
\hline
Absolutely no \{obj\} is here, yet \{cap\} remains. & Nowhere can \{obj\} be found, but \{cap\} is evident. \\
\hline
Nowhere in sight is any \{obj\}, yet \{cap\} unfolds. & No \{obj\} is around in the surroundings, but \{cap\} is occurring. \\
\hline
\end{tabular}
\caption{Templates for compositional negation caption generation, we replace the ${cap}$ with the provided original caption and ${obj}$ with the corresponding negation object.}
\label{table:compositional-template}
\end{table*}

\begin{table*}[h]
\centering
\begin{tabular}{|p{6cm}|p{5cm}|p{5cm}|}
\hline
\textbf{compositional negation caption} & \textbf{positive question} & \textbf{negative question}\\
\hline
A room painted in blue with a white sink, but no door. & Is there a room painted in blue with a white sink? & Is there a door? \\
\hline
A shot inside a kitchen without anyone present. & Is there a kitchen shown? & Is there anyone present? \\
\hline
A woman is walking on the sidewalk without her dog. & Is there a woman walking on the sidewalk? & Is there her dog? \\
\hline
A man without a bike at a marina. & Is there a man at a marina? & Is there a bike? \\
\hline
A man is sitting on a bench without a bicycle nearby. & Is there a man sitting on a bench? & Is there a bicycle nearby? \\
\hline
There's no kitchen sink beside the door and countertop. & Is there a door and countertop? & Is there a kitchen sink beside the door and countertop? \\
\hline
A bathroom without a checkered black and white tile floor. & Is there a bathroom? & Is there a checkered black and white tile floor? \\
\hline
A house boat is moored on a riverbank with no bikes in sight. & Is there a house boat moored on a riverbank? & Is there a bike? \\
\hline
A train missing a striped door waiting on a train track. & Is there a train waiting on a train track? &  Is there a striped door? \\
\hline
A small airplane flying without a jet nearby. & Is there a small airplane flying? & Is there a jet nearby? \\
\hline
A woman is seen without a horse in front of a fence with razor wire. & Is there a woman in front of a fence with razor wire? & Is there a horse? \\
\hline
No vans are traveling over a bridge next to train tracks. & Is there a bridge next to train tracks? & Is there a van? \\
\hline
A person riding a bicycle without any river nearby. & Is there a person riding a bicycle? & Is there a river nearby? \\
\hline
No giraffes can be seen in the wood and metal fenced enclosure. & Is there a wood and metal fenced enclosure? & Is there a giraffe? \\
\hline
A row team without a lead woman shouting. & Is there a row team? & Is there a lead woman shouting?\\
\hline
A lady is sitting in a room devoid of any bright pink walls. & Is there a lady sitting in a room? & Is there a bright pink wall?\\
\hline
A man carrying a plate without any food on it. & Is there a man carrying a plate? & Is there any food on the plate? \\
\hline

\end{tabular}
\caption{Example from \textit{Neg-TtoI} negation image generation benchmark}
\label{table:TtoI-example}
\end{table*}
\end{document}